\documentclass[10pt,twocolumn,letterpaper]{article}

\usepackage{cvpr}              % To produce the CAMERA-READY version
\usepackage{booktabs}
\usepackage{makecell}
\usepackage{array}
\usepackage{pgfplots}

\newcommand{\myparagraph}[1]{\noindent\textbf{#1.}}

\definecolor{cvprblue}{rgb}{0.21,0.49,0.74}
\usepackage[pagebackref,breaklinks,colorlinks,allcolors=cvprblue]{hyperref}
\usepackage{arydshln}

\def\paperID{****} % *** Enter the Paper ID here
\def\confName{CVPR}
\def\confYear{2026}

\newcommand{\methodname}{EgoFlow~} %tentative
\title{EgoFlow: Gradient-Guided Flow Matching \\ for Egocentric 6DoF Object Motion Generation}

\author{Abhishek Saroha$^{1,2}$
\quad
Huajian Zeng$^{4}$
\quad
Xingxing Zuo$^{4}$
\quad
Daniel Cremers$^{1,2}$
\quad
Xi Wang$^{1,2,3}$
\vspace{0.3em}\\  $^{1~}$TU München \qquad $^{2~}$MCML \qquad $^{3~}$ETH Zürich \qquad $^{4~}$MBZUAI
}
\begin{document}

\twocolumn[{%
\renewcommand\twocolumn[1][]{#1}%
\maketitle
\begin{center}%
    \centering%
    \vspace{-2cm}
    \captionsetup{type=figure}%
    \includegraphics[width=0.9\textwidth]{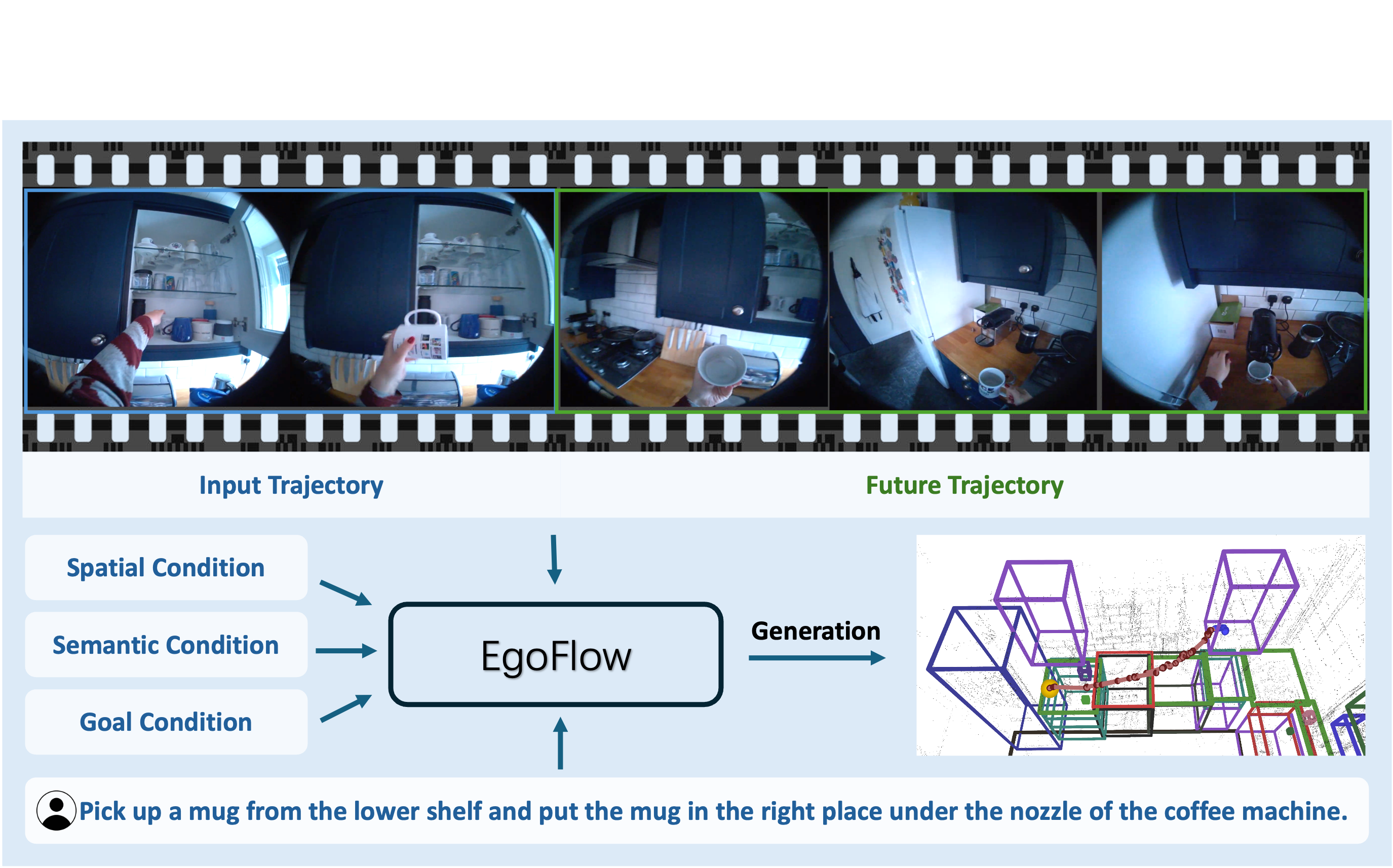}
    \caption{\textbf{EgoFlow: a method for object trajectory generation from egocentric videos.} Given a textural command and the surrounding environment, \methodname generates physically valid 6DOF object trajectories that respect spatial constraints across diverse environments by learning from egocentric videos. 
    }
    \label{fig:teaser}
\end{center}%
}]

\maketitle
\begin{abstract}
Understanding and predicting object motion from egocentric video is fundamental to embodied perception and interaction. However, generating physically consistent 6DoF trajectories remains challenging due to occlusions, fast motion, and the lack of explicit physical reasoning in existing generative models. We present EgoFlow, a flow-matching framework %on SE(3) \as{or 9D euclidean representation?} 
that synthesizes realistic and physically plausible trajectories conditioned on multimodal egocentric observations. \methodname employs a hybrid Mamba–Transformer–Perceiver architecture to jointly model temporal dynamics, scene geometry, and semantic intent, while a gradient-guided inference process enforces differentiable physical constraints such as collision avoidance and motion smoothness. This combination yields coherent and controllable motion generation without post-hoc filtering or additional supervision. Experiments on real-world datasets HD-EPIC, EgoExo4D, and HOT3D show that \methodname outperforms diffusion-based and transformer baselines in accuracy, generalization, and physical realism, reducing collision rates by up to 79\%, and strong generalization to unseen scenes. Our results highlight the promise of flow-based generative modeling for scalable and physically grounded egocentric motion understanding. Project page: \url{https://abhi-rf.github.io/egoflow/}.

%Egocentric datasets provide diverse observations of object motion across real-world scenes. We generate 6-DoF trajectories using flow matching on SE(3), which learns motion distributions from demonstrations via continuous normalizing flows. While flow matching produces natural motion patterns, it cannot guarantee physical validity as models trained solely on visual examples cannot always reason about constraints in novel environments. We address this limitation through gradient-guided sampling that optimizes generated trajectories using differentiable physical cost functions, enforcing collision avoidance and motion smoothness. We empirically validate our approach on real-world egocentric datasets, demonstrating significant improvements in collision avoidance and cross-environment generalization compared to state-of-the-art trajectory generation methods, generating trajectories that could potentially be useful in further downstream tasks. 

\end{abstract}

\vspace{-0.6cm}    
\section{Introduction}
\label{sec:intro}

Recent advances in augmented reality devices~\cite{engel2023project} have enabled large-scale egocentric datasets with fine-grained spatial and semantic annotations~\cite{raina2023aria, Pan_2023_ICCV,lv2024aria, perrett2025hd, grauman2024ego, banerjee2025hot3d}, providing new opportunities for data-driven learning of object motion trajectories in realistic environments.
These egocentric videos capture a rich source of information about object interactions, providing valuable cues about how objects move, are interacted, and evolve over time. 
Such data offers dense, first-person evidence for embodied perception and interaction in robotics, where understanding and predicting object dynamics is essential for planning and interaction.
% ~\cite{zeng2026flowhoi}. %looks out of place here.

% challanges
Object motion generation models the temporal evolution of object movements, revealing how humans interact with their surroundings from a first-person viewpoint. 
However, learning to generate object trajectories from these egocentric videos poses unique challenges.
First, egocentric scenes are highly diverse and cluttered, with objects frequently occluded. In addition, egocentric videos have limited field of view and rapid camera motions lead to blurring content. These factors require models to reason over complex geometric layouts and semantic cues to generate accurate trajectories. 
Second, long-horizon prediction demands consistent temporal reasoning, as small spatial errors can accumulate and lead to unrealistic motion patterns over time.
Third, ensuring physical plausibility without explicit physics supervision is challenging; generated trajectories must remain collision-free and dynamically smooth while reflecting realistic motion patterns.
%
% Unlike static scene analysis, 
% However, learning from egocentric videos poses unique challenges for machine perception due to frequent occlusions, rapid viewpoint changes, and limited field of view. 
% Furthermore, generating realistic consistent 6DoF trajectories requires explicit physical reasoning.

% how to solve it
To address these challenges, we propose a generative model \methodname that learns scene-conditioned continuous flows for 6DoF object trajectory generation. Rather than using stochastic diffusion, our method leverages flow matching~\cite{lipman2022flow, liu2022flow} to learn deterministic transport fields in $\mathbb{R}^9$, enabling efficient trajectory synthesis. Building upon the multimodal conditioning framework introduced in GMT~\cite{zeng2026gmt}, \methodname further incorporates a hybrid architecture that combines bidirectional Mamba state-space models~\cite{gu2024mamba, zhang2024motion} with Transformer blocks and a Perceiver-based cross-attention encoder~\cite{jaegle2021perceiver}, allowing scalable sequence modeling and effective fusion of geometric, semantic, and goal-oriented context. Finally, we introduce a gradient-guided sampling strategy that injects differentiable physical costs, specifically SDF-based collision penalties, rotation and velocity smoothness, into the generation loop to improve physical plausibility without requiring explicit constraint supervision.

% briefly introduce how we evalute it
We evaluate \methodname on large-scale egocentric benchmarks to verify both fidelity and generalization. Specifically, we train and test on HD-EPIC~\cite{perrett2025hd} to assess physical consistency under realistic scene contexts. Following a similar protocol as in previous work~\cite{yoshida2025generating}, we perform zero-shot transfer from Ego-Exo4D~\cite{grauman2024ego} to HOT3D~\cite{banerjee2025hot3d} to evaluate robustness in unseen environments. \methodname achieves substantial gains over existing generative models, %including GIMO~\cite{zheng2022gimo}, CHOIS~\cite{li2024controllable}, and EgoScaler~\cite{yoshida2025generating}, 
particularly in terms of spatial accuracy and physical feasibility, reducing collision rates by up to 79\%. 
\methodname also shows strong generalization to unseen scenes. Our code is publicly available to facilitate future research.

In summary, our contributions are:
\begin{itemize}
      \item A scene-conditioned flow matching method for 6DoF object motion generation, which learns continuous flows in Euclidean space $\mathbb{R}^9$ for efficient and physically consistent motion synthesis.
      \item A hybrid architecture that integrates bidirectional Mamba state-space models with Transformer and Perceiver components, enabling scalable sequence modeling and effective fusion of geometric and semantic scene features.
      \item A gradient-guided sampling strategy that incorporates differentiable physical costs into the generation process, enhancing the physical plausibility of generated trajectories without explicit supervision.
      \item Comprehensive experiments on egocentric datasets, demonstrating substantial improvements over existing  models and strong generalization to unseen scenes.
\end{itemize}

\begin{figure*}[t!]
    \centering
    \includegraphics[width=0.95\textwidth]{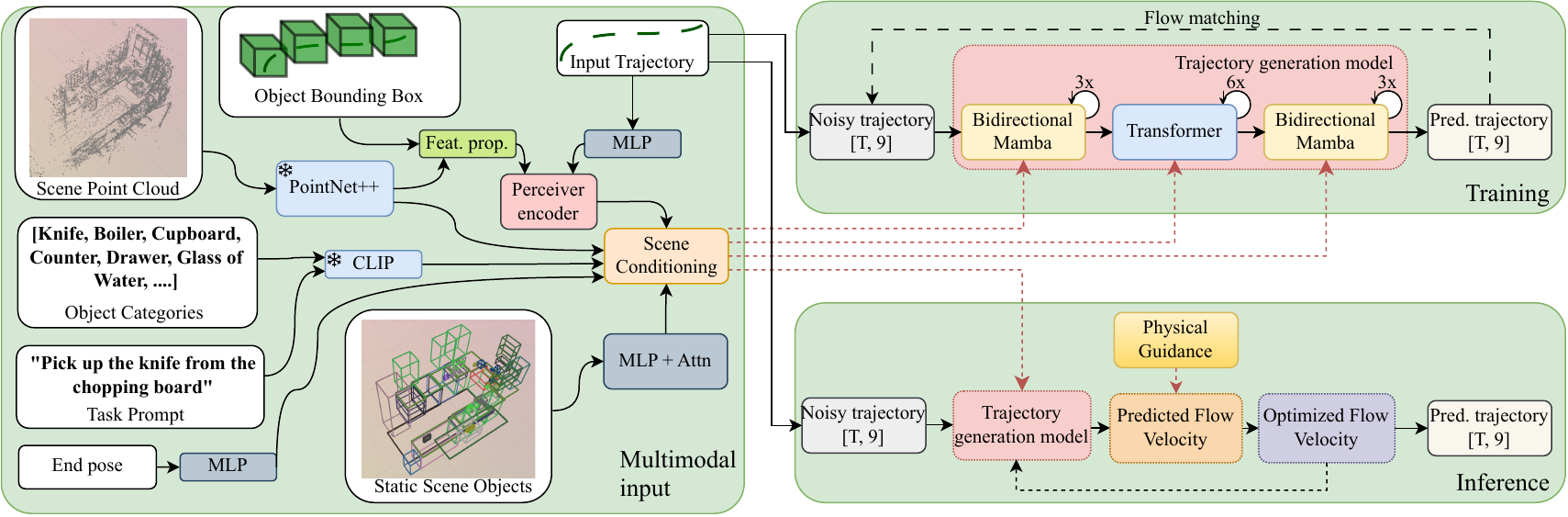}
    \caption{
    \textbf{\methodname overview.} Given a 3D scene, a task prompt, and a task goal, our method first fuses multimodal inputs through a scene conditioning block (Sec.~\ref{sec:conditioning}). The fused features are used as conditioning for trajectory generation. We use input trajectories as the source samples of our flow matching model (Sec.~\ref{subsec:flow_matching}), which maps the generated trajectories to the target distribution, the ground-truth trajectories, through a hybrid architecture (Sec.~\ref{subsec:architecture}). We integrate physical guidance at inference to ensure physical plausible and collision-free trajectories (Sec.~\ref{sec:guidance}).      
    }
    \label{fig:method}
    \vspace{-0.5cm}
\end{figure*}

\section{Related Work}
\label{sec:related}

\myparagraph{Motion and Trajectory Generation}
Modeling 3D motion from visual or multimodal data has been widely studied across human motion~\cite{guo2022generating,tevet2022human,zhang2023generating,guo2024momask}, object dynamics~\cite{yoshida2025generating,li2024controllable}, and scene-conditioned motion prediction~\cite{zheng2022gimo}. 
Early approaches often relied on short-term kinematic priors or autoregressive prediction, which limited their ability to capture long-range dependencies and global temporal coherence.
Recent progress in diffusion-based generative modeling has enabled high-fidelity motion synthesis conditioned on language, geometry, or scene context~\cite{li2024controllable,zhang2024hoidiffusion,wu2025human}, showing that large-scale motion datasets can be effectively modeled as continuous spatiotemporal distributions.
However, existing methods remain constrained to short temporal horizons and often overlook explicit physical consistency such as collision avoidance or trajectory smoothness. In contrast, our framework focuses on generating long-horizon, physically consistent object trajectories.

%%%%%%%%%%%%%%%%%%
\myparagraph{Egocentric Motion Understanding}
Egocentric perception provides a natural framework for studying object and scene dynamics from a first-person viewpoint.
Large-scale benchmarks such as EgoExo4D~\cite{grauman2024ego}, HOT3D~\cite{banerjee2025hot3d}, and HD-EPIC~\cite{perrett2025hd} enable fine-grained analysis of 3D trajectories, spatial context, and visual cues observed during everyday activities. 
These datasets have motivated new research in egocentric motion forecasting, trajectory synthesis, and cross-view understanding, where models learn to infer plausible 3D trajectories directly from wearable sensors or video streams.
For example, EgoScaler~\cite{yoshida2025generating} generates object trajectories from egocentric visual inputs, while EgoChoir~\cite{yang2024egochoir} explores scene affordance reasoning from egocentric observations.
Despite their success, most existing approaches focus on short-term motion generation or limited geometric reasoning. 
Our work advances this line of research by introducing a continuous flow generative model capable of synthesizing long-horizon, physically consistent trajectories conditioned on egocentric scene context.
%%%%%%%%%%%%%%%%%%%

\myparagraph{Generative Models and Mamba}
Diffusion-based generative models~\cite{ho2020denoising,song2019generative,song2020score} have established a new foundation for high-quality synthesis across images, audio, and motion.
Their continuous extensions using stochastic differential equations provide a principled view of data generation as iterative refinement along a noise-to-data path.
More recently, flow matching~\cite{lipman2022flow,tong2023improving} has emerged as a deterministic alternative, learning continuous velocity fields that transport samples from noise to the target manifold.
This formulation offers comparable expressiveness while enabling faster and more stable inference.

Parallel to these algorithmic advances, architectural innovations have improved the scalability of sequence modeling.
The Mamba architecture~\cite{gu2024mamba,dao2024transformers} introduces a selective state-space mechanism~\cite{gu2021combining,gu2021efficiently,smith2022simplified} that supports linear-time inference and constant memory cost, making it suitable for very long sequences.
Variants such as Vision Mamba~\cite{zhu2024vision} and DiM~\cite{teng2024dim,fei2024dimba} demonstrate strong performance in large-scale visual and generative tasks.

Building on these advances, our framework integrates flow matching with Mamba-based sequence modeling to achieve scalable, coherent, and physically consistent trajectory generation from complex egocentric scenes.

\noindent\textbf{Concurrent Work.}
ObjectForesight~\cite{soraki2026objectforesightpredictingfuture3d} predicts future 6DoF object trajectories from egocentric video using diffusion. FlowHOI~\cite{zeng2026flowhoi} generates hand-object interaction sequences via two-stage flow matching conditioned on egocentric observations and 3D scene reconstructions. EgoMAN~\cite{chen2025flowingreasoningmotionlearning} and EgoVerse~\cite{punamiya2026egoverse} provide large-scale egocentric datasets with 6DoF hand trajectories and diverse manipulation demonstrations, complementary to our work. GMT~\cite{zeng2026gmt} addresses goal-conditioned 6DoF object trajectory synthesis using a multimodal transformer and shares the input scene conditioning design with our method.

\section{Methodology}
\label{sec:method}

Object trajectory generation requires capturing both the diversity of feasible motions and the physical regularities that govern real-world interactions.
To address this dual challenge, the proposed framework formulates trajectory synthesis as a continuous transport process in $\mathbb{R}^{9}$, where each frame is represented as position 
$\mathbb{R}^{3}$ and continuous 6D rotation~\cite{zhou2019continuity}. Motion evolution is, therefore, learned as a deterministic flow field rather than a stochastic diffusion.
A hybrid Mamba-Transformer backbone further provides structured temporal reasoning and multimodal scene understanding, supporting long-horizon prediction under complex spatial and semantic constraints.
Physical consistency during inference is ensured through gradient-guided optimization that dynamically aligns generated trajectories with collision and stability constraints.
Together, these components form a unified generative framework capable of producing smooth, physically grounded 6DoF trajectories (see Fig.~\ref{fig:method}).

% \vspace{-0.9em}
\subsection{Problem Formulation}
\label{subsec:problem}
We begin by formally defining the object trajectory generation problem.
Given a 3D scene with known spatial and semantic context, we aim to generate physically plausible 6DoF object trajectories that satisfy motion and collision constraints. Each trajectory frame is represented as $\mathbf{x}_t = [\mathbf{p}_t; \mathbf{r}_t] \in \mathbb{R}^9$ where $\mathbf{p}_t \in \mathbb{R}^3$ is the 3D position and $\mathbf{r}_t \in \mathbb{R}^6$ is the continuous 6D rotation representation~\cite{zhou2019continuity}.
The input conditions include a point cloud of the scene $\mathcal`{P} \in \mathbb{R}^{N \times 3}$, $M$ oriented bounding boxes of fixtures $\mathcal{B} = \{(\mathbf{c}_i, \mathbf{s}_i, \mathbf{r}_i)\}_{i=1}^{M}$ representing centers, sizes, and rotations, object category $\mathcal{C}$, target end pose $\mathbf{x}_T \in \mathbb{R}^9$ for goal conditioning, text prompt describing the task, and observed trajectory history $\mathbf{x}_{1:H} \in \mathbb{R}^{H \times 9}$. 
Using a radius based selection, the number of fixatures $M$ is typically around 50. 
The goal is to generate future trajectory $\mathbf{x}_{H+1:T} \in \mathbb{R}^{(T-H) \times 9}$ that captures plausible object motion while avoiding collisions.
In practice, we consider the $30\%$ of the trajectory as input and predict the remaining $70\%$.

%%%%%%%%%%%%%%%%%%%%%%%%%%%%%%%%%%%%%%%%%%%%%%%%%%%%%%%%%%%%
\subsection{Multimodal Conditioning}
\label{sec:conditioning}
% \noindent\textbf{Multimodal Conditioning.}
% Before processing the trajectory sequence, 
Following the multimodal conditioning design of GMT~\cite{zeng2026gmt}, we first build a unified conditioning representation $\mathbf{u}$ by encoding the following multimodal input.

\noindent\emph{(i) Trajectory dynamics.}
The observed history $\mathbf{x}_{1:H}$ is linearly projected to a temporal embedding $\mathbf{F}_{\text{traj}}$ that preserves local kinematics while remaining compatible with cross-modal fusion.

\noindent\emph{(ii) Local geometric context.}
The raw scene point cloud $\mathcal{P}\in\mathbb{R}^{N\times 3}$ is encoded by PointNet++~\cite{qi2017pointnet++} to produce a global descriptor $\mathbf{F}_{g}$ and per-point features $\mathbf{F}_{p}$. To provide geometry precisely “at” the manipulated object at each time $t$, per-point features are propagated from nearby points to the object's oriented box center $\mathbf{c}_t$ using inverse-distance weighting over the $k$ nearest neighbors:
\begin{equation}
\mathbf{F}{p} = \sum_{t=1}^H \frac{\sum_{i=1}^k w_i(\mathbf{c}_t)\mathbf{f}_i}{\sum_{i=1}^k w_i(\mathbf{c}_t)}, \quad
w_i(\mathbf{c}_t) = \frac{1}{|\mathbf{c}_t - \mathbf{p}_i|^2},
\end{equation}
where $\mathbf{p}_i$ and $\mathbf{f}_i$ are the coordinates and features of the $i$-th neighbor point.

\noindent\emph{(iii) Fixture layout.}
Static fixtures $\mathcal{B}$ are embedded as tokens from box geometry $(\mathbf{c}_i,\mathbf{s}_i,\mathbf{r}_i)$. Geometric tokens interact via a lightweight self-attention layer to capture pairwise relations (e.g., “counter above cabinet”). To reduce noise, only the $M$ nearest fixtures to the object are retained before attention:
\begin{equation}
    \mathbf{F}_b = \mathrm{SelfAttn}\big(\{b_k\}_{k=1}^M\big).
\end{equation}

\noindent\emph{(iv) Category and task prompt.}
Object category $\mathcal{C}$ and task description are embedded by CLIP and projected by MLP to $\mathbf{F}_{s}$, providing behavior priors that complement purely geometric cues (e.g., “drawer” vs. “box” implies different feasible orientations).

\noindent\emph{(v) Goal descriptor.}
The target pose $\mathbf{x}_T\in\mathbb{R}^9$ is mapped by MLP to a goal token $\mathbf{F}_{goal}$ that conditions long-horizon planning and mitigates drift from the intended destination.

All conditioning features are concatenated and projected to form the final conditioning vector $\mathbf{u}$:
\begin{equation}
\mathbf{u} = \text{MLP}([\mathbf{F}_{\text{traj}}, \mathbf{F}_{p}, \mathbf{F}_{g}, \mathbf{F}_{b}, \mathbf{F}_{s}, \mathbf{F}_{\text{goal}}]).
\end{equation}
This unified conditioning vector is shared across all model stages but modulated differently at each stage to enable hierarchical fusion of spatial, semantic, and task-specific cues.

%%%%%%%%%%%%%%%%%%%%%%%%%%%%%%%%%%%%%%%%%%%%%%%%%%%%%%%%%%%%
\subsection{Flow Matching for Trajectory Generation}
\label{subsec:flow_matching}

The trajectory distribution is modeled using flow matching~\cite{lipman2022flow, liu2022flow}, a class of continuous normalizing flows that learn deterministic data transport between probability distributions.
In contrast to diffusion-based models, flow matching offers two distinct advantages:
First, it learns direct velocity fields through simple regression without requiring stochastic differential equation solvers or score estimation.
Second, it produces smoother probability paths, enabling efficient generation with fewer integration steps.
We extend this formulation to handle 6DoF trajectories conditioned on history observation.

\noindent\textbf{Training.} Given a clean trajectory $\mathbf{x}_0 \sim p_{\text{data}}$ and noise $\mathbf{x}_1 \sim \mathcal{N}(0, \mathbf{I})$, we define a linear interpolation:
\begin{equation}
\mathbf{x}_t = (1-t)\mathbf{x}_0 + t\mathbf{x}_1, t \in [0,1].
\end{equation}
The model learns a time dependent velocity field $\mathbf{v}_\theta$ conditioned on the scene context $\mathcal{S}$:
\begin{equation}
\mathcal{L}_{\text{FM}} = 
\mathbb{E}_{t \sim \mathrm{U}[0,1],\, \mathbf{x}_0, \mathbf{x}_1}
\left[\|\mathbf{v}_\theta(\mathbf{x}_t, t, \mathcal{S}) - (\mathbf{x}_1 - \mathbf{x}_0)\|_1\right],
\end{equation}
where $\mathbf{v}_\theta$ predicts the instantaneous flow direction at intermediate states.
We apply equal L1 loss weights to the position and rotation components of 6DoF trajectory, treating the continuous 6D rotation~\cite{zhou2019continuity} as an embedding in $\mathbb{R}^6$.

\noindent\textbf{Sampling} At inference, we draw an initial sample $\mathbf{x}_1 \sim \mathcal{N}(0,\mathbf{I})$ and integrate the learned velocity field backward using Euler's method~\cite{robinson2004introduction}:
\begin{equation}
\mathbf{x}_{t-\Delta t} = \mathbf{x}_t - \Delta t \cdot \mathbf{v}\theta(\mathbf{x}_t, t, \mathcal{S}),
\end{equation}
with $\Delta t = 1/20$ for 20 steps.
This deterministic integration produces smooth and physically consistent trajectories within the learned manifold.

%%%%%%%%%%%%%%%%%%%%%%%%%%%%%%%%%%%%%%%%%%%%%%%%%%%%%%%%%%%%

\subsection{Hybrid Architecture}
\label{subsec:architecture}
A hybrid architecture is used to combines efficient sequence modeling with expressive multimodal reasoning to support long-horizon trajectory generation under complex scene constraints. The model integrates three stages: temporal encoding, cross-modal reasoning, and trajectory refinement. %contributing to progressively structured motion prediction.

\noindent\textbf{Stage 1: Temporal Context Encoding.}
The first stage employs three bidirectional Mamba layers~\cite{zhang2024motion} to capture long range temporal dependencies with linear computational complexity.
Each layer updates the trajectory embedding $\mathbf{h}_t$ through feature-wise linear modulation (FiLM)~\cite{perez2018film}: $\mathbf{h}_t' = \gamma(\mathbf{u}_t) \odot \mathbf{h}_t + \beta(\mathbf{u}_t),$
where $\odot$ denotes element-wise (Hadamard) multiplication, and $\gamma$ and $\beta$ are learnable affine transformations of the conditioning vector.
Bidirectional processing combines forward and backward Mamba passes, enabling the model to reason over both past and future temporal context, crucial for generating collision-free, dynamically consistent motion.

\noindent\textbf{Stage 2: Cross-Modal Attention.}
The second stage consists of six Transformer blocks adapt from the Perceiver architecture~\cite{jaegle2021perceiver}. First, self-attention captures temporal dependencies among trajectory tokens. Next, cross-attention allows trajectory features to query the full multimodal conditioning $\mathbf{u}_t$, enabling selective integration of relevant scene and task information. Finally, a position-wise feedforward network (FFN) refines the fused representation. Note that unlike FiLM's affine modulation, cross-attention provides learned attention weights that enable the model to focus on pertinent conditioning information (e.g., nearby obstacles, goal location) for each trajectory segment, enhancing multimodal reasoning.

\noindent\textbf{Stage 3: Trajectory Refinement.}
The final stage applies another stack of three bidirectional Mamba layers, again conditioned via FiLM.
This stage refines the fused representation from Stage 2, integrating temporal coherence with the enriched multimodal context. The output is projected to the $\mathbb{R}^{9}$ velocity space, comprising both linear and angular velocity components via a linear head, producing the final velocity prediction $\mathbf{v}_\theta(\mathbf{x}_t, t, \mathcal{S})$ for flow matching.

%%%%%%%%%%%%%%%%%%%%%%%%%%%%%%%%%%%%%%%%%%%%%%%%%%%%%%%%%%%%

% \subsection{Gradient-Guided Sampling for Physical Constraint Satisfaction}
\subsection{Gradient-Guided Sampling}
\label{sec:guidance}

Models trained solely on collision-free demonstrations cannot explicitly reason about physical constraints, failing to avoid obstacles in novel scene configurations at test time. We address this through gradient-guided
sampling~\cite{janner2022planning,huang2023diffusion,yan2025m} that enforces constraints via differentiable optimization during inference. This approach avoids train-test distribution mismatch: the flow matching model learns motion
distributions from data, while physical constraints are imposed only during generation when needed.

At each integration timestep $t$, before computing $\mathbf{x}_{t-\Delta t} = \mathbf{x}_t - \Delta t \cdot \mathbf{v}_\theta$, we refine the predicted velocity through $K{=}50$ gradient descent steps with learning rate $\alpha{=}0.1$ and in each step $k$: 
\begin{equation}
\mathbf{v}_{\theta}^{(k+1)} = \mathbf{v}_{\theta}^{(k)} - \alpha \nabla_{\mathbf{v}} \mathcal{J}(\mathbf{x}_t - \Delta t \cdot \mathbf{v}^{(k)}),
\end{equation}
where $\mathcal{J}$ combines three differentiable costs computed on the integrated trajectory state.

\noindent\textbf{Collision avoidance.} We compute signed distance fields between trajectory positions and static fixtures $\mathcal{B}$ using analytic formulas for oriented bounding boxes~\cite{quilez2008sdf}. The minimum distance from the position $\mathbf{p}_j$ to all $M$ fixtures is:
\begin{equation}
d(\mathbf{p}_j) = \min_{i=1}^M \text{SDF}(\mathbf{p}_j, \mathbf{c}_i, \mathbf{s}_i, \mathbf{r}_i).
\end{equation}
The cost penalizes violations of a safety margin $\epsilon{=}5$cm:
\begin{equation}
\mathcal{J}_{\text{coll}} = \sum_{j=H+1}^T \max(0, \epsilon - d(\mathbf{p}_j)).
\end{equation}

\noindent\textbf{Rotational consistency.} Physically plausible trajectories require smooth angular motion with consistent velocity directions, which is enforced in the rotation space by measuring cosine similarity between consecutive
rotation changes:
\begin{equation}
\mathcal{J}_{\text{rot}} = \sum_{j=H+1}^{T-1} \left(1 - \frac{\langle \Delta \mathbf{r}_j, \Delta \mathbf{r}_{j-1} \rangle}{\|\Delta \mathbf{r}_j\| \|\Delta \mathbf{r}_{j-1}\|}\right), \quad \Delta \mathbf{r}_j = \mathbf{r}_{j+1} -
\mathbf{r}_j,
\end{equation}
which penalizes abrupt changes in rotation direction.

\noindent\textbf{Translational smoothness.} We enforce smooth velocity profiles by penalizing linear acceleration, which prevents abrupt velocity changes as per the following:
\begin{equation}
\mathcal{J}_{\text{vel}} = \sum_{j=H+1}^{T-2} \|\mathbf{a}_j\|, \quad \mathbf{a}_j = \mathbf{v}_{j+1} - \mathbf{v}_j, \quad \mathbf{v}_j = \frac{\mathbf{p}_{j+1} - \mathbf{p}_j}{\Delta t}.
\end{equation}

The total cost is $\mathcal{J} = \mathcal{J}_{\text{coll}} + \lambda_{\text{rot}} \mathcal{J}_{\text{rot}} + \lambda_{\text{vel}} \mathcal{J}_{\text{vel}}$ and its full differentiability enables gradient-based refinement toward physically plausible trajectories. Further analysis on the formulation and complexity of the gradient-guided sampling can be found in the Supp. Mat. 
%%%%%%%%%%%%%%%%%%%%%%%%
\subsection{Implementation Details}

\noindent\textbf{Training.} We train using AdamW with learning rate $10^{-4}$, batch size 32, for 100 epochs. The implementation uses PyTorch Lightning for efficiency.

\noindent\textbf{Inference.} Gradient guidance applies 50 gradient descent steps per integration step with learning rate $\alpha = 0.1$ and collision margin $\epsilon = 0.05$m. Smoothness weights are tuned per dataset: $\lambda_{\text{rot}} = \lambda_{\text{vel}} = 2.0$ for HD-EPIC, $\lambda_{\text{rot}} = \lambda_{\text{vel}} = 5.0$ for HOT3D. Architectural hyperparameters (layer counts, hidden dimensions, attention heads) are detailed in the supplementary material.

\begin{figure*}[h!]
    \centering
    \small
    \includegraphics[width=1\textwidth]{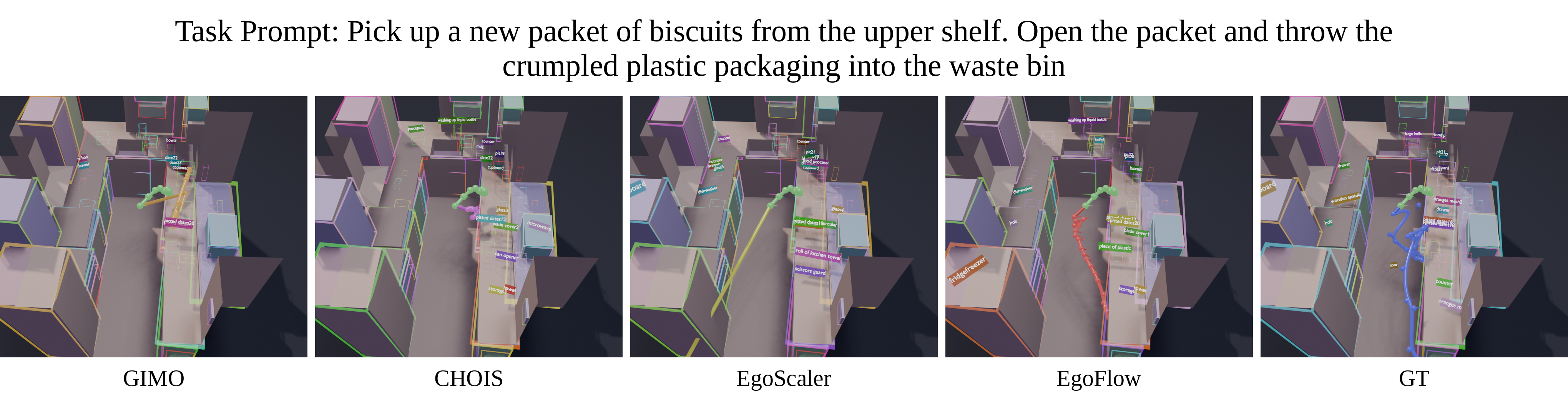}
    \vspace{-0.86cm}
    \caption{
    \textbf{HD-Epic Qualitative Result.} The trajectory in green in each image is the history followed by the respective prediction by the various baselines and the ground truth. We can see that not ony our method generates a plausible trajectory to the end goal, it also takes a rather more natural and smooth path to the target pose. 
    }
    \label{fig:hdepic_qualitative}
    \vspace{-0.2cm}
\end{figure*}

\begin{figure*}[t]
    \centering
    \small
    \includegraphics[width=0.95\textwidth]{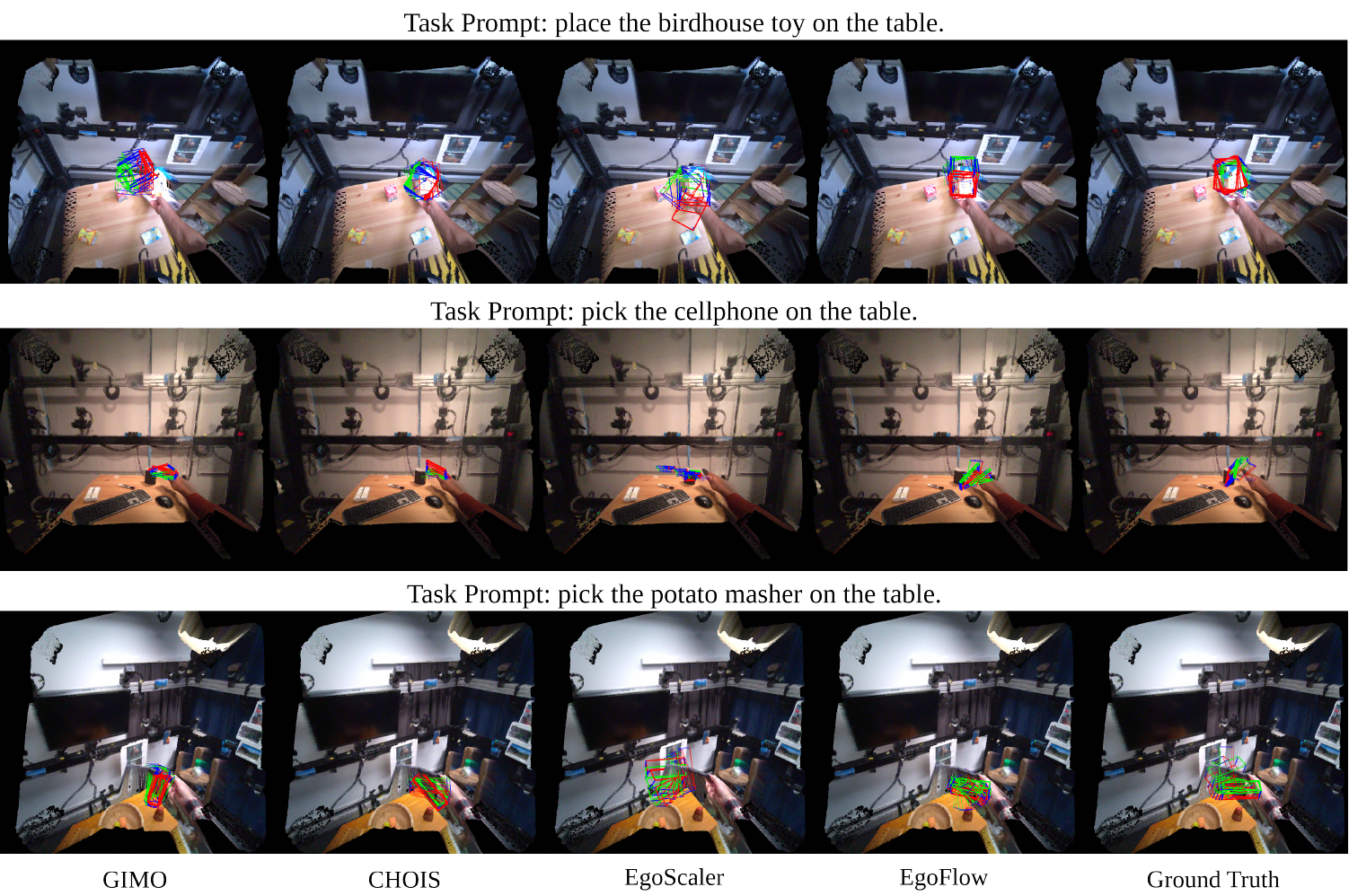}
    \vspace{-0.35cm}
    \caption{
    \textbf{Hot3D Qualitative Results.} We compare \methodname against the established baselines. Our method shows better generalization to unseen conditions and produce geometrically coherent and physically plausible 6DoF trajectories.
    }
    \label{fig:hot3d_qualitative}
        \vspace{-0.4cm}
\end{figure*}

\section{Experiments}
\label{sec:exp}

% short outline
%In this section, 
We present the evaluation of our proposed method by first describing baselines and evaluation metrics used in our experiments. Next, we compare our method with baselines on various benchmarks. Finally, we analyze the results and discuss the implications of our findings.
% \todo{need more specific information and add more details here}

% Our goal is to examine the following questions:

% \begin{itemize}

%     \item How does our method perform compared to existing state-of-the-art approaches in terms of accuracy and robustness?
%     \item Which scenarios or conditions does our method excel in, and where does it face challenges?
%     \item What is the contribution of each part of our design to the final performance? Overall, is the proposed architecture effective?

% \end{itemize}
\subsection{Experiments Setup}
\label{subsec:setup}

\noindent\textbf{Baselines.} Given our task setting, there are no directly comparable existing methods. Following the baseline adaptation protocol of GMT~\cite{zeng2026gmt}, we select three relevant approaches and adapt them to our setting, and additionally evaluate four recent diffusion/flow-based methods retrained on our data:
\begin{itemize}
\item \textbf{EgoScaler}~\cite{yoshida2025generating}: A vision-language-based generative framework that synthesizes 6DoF object manipulation trajectories from textual action descriptions and egocentric visual inputs. We adopt their \textbf{PointLLM}~\cite{xu2024pointllm} variant and retrain under our dataset configuration.

\item \textbf{GIMO}~\cite{zheng2022gimo}: A Perceiver-based transformer for egocentric human motion forecasting. We replace the human body representation with our object-centric descriptor $\mathbf{x}_t = [\mathbf{p}_t; \mathbf{r}_t] \in \mathbb{R}^9$ and add goal conditioning.

\item \textbf{CHOIS}~\cite{li2024controllable}: A generative framework for human-object interaction conditioned on object geometry, waypoints, and text. We retain only the object trajectory branch and modify the waypoint conditioning to first 30\% of input.

\item \textbf{M2Diffuser}~\cite{yan2025m2diffuser}: A diffusion-based model that generates scene-conditioned whole-body mobile manipulation trajectories from 3D scans, with differentiable physical cost functions enforced during inference.

\item \textbf{DP3}~\cite{Ze2024dp3}: A visuomotor imitation learning method that generates robot actions via diffusion conditioned on compact 3D representations extracted from point clouds.

\item \textbf{SPOT}~\cite{hsu2025spotse3posetrajectory}: An object-centric framework that first generates SE(3) object pose trajectories via diffusion, then uses them to condition a robot action policy, enabling cross-embodiment transfer.

\item \textbf{ManiFlow}~\cite{Yan2025ManiFlow}: A visuomotor imitation learning policy that combines flow matching with consistency training for efficient action generation from visual, language, and proprioceptive inputs.

\end{itemize}

\noindent\textbf{Evaluation Metrics.}
To quantitatively evaluate the quality of the predicted object trajectories, we employ a set of metrics designed to capture positional accuracy, temporal coherence, and physical realism:
\begin{itemize}

    \item \textbf{Average Displacement Error (ADE)}: Computes the mean Euclidean (L2) distance (in metres) between predicted and ground-truth positions over all future time steps, reflecting the overall accuracy of the prediction.

    \item \textbf{Final Displacement Error (FDE)}: Measures the Euclidean distance (in metres) between the predicted and ground-truth positions at the last prediction step, emphasizing the accuracy of the final position.

    \item \textbf{Fréchet Distance}~\cite{efrat2002new}: Quantifies the maximum deviation between two trajectories while accounting for optimal temporal alignment. This metric jointly evaluates spatial and temporal consistency. Smaller values indicate closer adherence of the predicted trajectory to the ground truth in both shape and timing.

    % \item \textbf{Angular Consistency}: Measures how well the directional dynamics of the predicted trajectory align with the reference sequence. The positional differences between consecutive frames are represented as direction vectors, and the mean cosine similarity between these vectors quantifies the preservation of orientation trends and motion smoothness. Higher values indicate better temporal coherence and directional stability. 

    \item \textbf{Geodesic Distance (GD)}~\cite{wohlhart2015learning}: Quantifies the instantaneous angular deviation between two rotations in radians, providing a measure of rotational error in 3D space.

    \item \textbf{Collision Rate}: Represents the proportion of trajectories intersecting with static scene elements, determined via bounding box overlap. Lower values suggest more physically plausible and spatially consistent predictions.

\end{itemize}

\begin{table}[t]
\centering
\small
\caption{\textbf{Quantitative Results: HD-EPIC} We compare model performance on various metrics on the HD-EPIC dataset. We can observe that our method performs the best against the baselines, while adding guidance sampling significantly reduces its collision rate. The results are averaged over 3 runs for EgoFlow.}
\label{tab:hdepic}
\vspace{-0.2cm}
\setlength{\tabcolsep}{3.6pt}
\begin{tabular}{p{1.8cm}ccccc}
\toprule
\textbf{Model} & \textbf{ADE$\downarrow$} & \textbf{FDE$\downarrow$} & \textbf{Frechet$\downarrow$} & \textbf{Geodesic$\downarrow$} & \textbf{Coll.$\downarrow$} \\
\midrule
GIMO~\cite{zheng2022gimo} & \underline{0.285} & 0.509 & \underline{0.210} & \textbf{0.725} & 23.5\% \\
CHOIS~\cite{li2024controllable} & 0.471 & 0.755 & 0.262 & 1.255 & 18.7\% \\
Egoscaler~\cite{yoshida2025generating} & 1.330 & 1.494 & 0.315 & 1.614 & 35.8\% \\
ManiFlow~\cite{Yan2025ManiFlow}& 1.214 & 1.573  & 0.609 & 1.692 &  35.5\% \\
DP3~\cite{Ze2024dp3} & 1.298 & 1.713  & 0.460 & 1.734  & 28.1\% \\
SPOT~\cite{hsu2025spotse3posetrajectory} & 1.440 & 1.795  & 0.469 & 1.740  & 28.7\% \\
M2Diffuser~\cite{yan2025m2diffuser} & 0.601 & \underline{0.442} & 0.476 & 1.788  & \underline{8.5\%}  \\
% \methodname w/o Guidance& \textbf{0.277} & \textbf{0.098} & \textbf{0.197} & 1.141 & \underline{11.6\%} \\
\methodname & \textbf{0.279} & \textbf{0.102} & \textbf{0.197} & \underline{1.141} & \textbf{2.5\%} \\
 % ~~w/o Guid. & \textbf{0.277} & \textbf{0.098} & \textbf{0.197} & 1.141 & \underline{11.6\%} \\
\bottomrule
\end{tabular}
\vspace{-0.5cm}
\end{table}

% \todo{change if evaluation metrics are different, add formula for frechet, angular, gd if space allows}
%%%%%%%%%%%%%%%%%%%%%
\subsection{Realistic Environments}
\label{subsec:hdepic}
\noindent\textbf{Dataset.}
We evaluate our framework on the HD-EPIC dataset~\cite{perrett2025hd}, which provides 41 hours of egocentric recordings captured across nine household kitchens using Project Aria glasses, accompanied by 3D digital twins, dense narrations, and temporally aligned action annotations. However, HD-EPIC offers only sparse object-level annotations at pickup and placement events, making it unsuitable for direct use in continuous trajectory generation.

\noindent\textbf{Preprocessing.}
We reconstruct continuous 6DoF object trajectories from the sparse annotations by leveraging the interacting hand as a physical proxy. Under the rigid-body coupling assumption during manipulation, we extract 6DoF hand trajectories via Project Aria's Machine Perception Service (MPS)~\cite{engel2023project} and transfer them to the object reference frame. Hand identity is initialized by proximity to the annotated object position and propagated using contact predictions from Hands23~\cite{cheng2023hands23}, with a sliding-window filter to ensure temporal consistency. Hand orientation is recovered via a 6D rotation representation derived from the wrist-palm coordinate frame using SVD. We refer readers to GMT~\cite{zeng2026gmt} and the Supp.\ Mat.\ for full preprocessing details.
\noindent\textbf{Results.}
Tab.~\ref{tab:hdepic} reports quantitative comparisons on HD-EPIC. Our approach outperforms all baselines across most metrics, demonstrating both spatial and physical consistency. Notably, our model achieves a collision rate of only 2.5\%, a substantial improvement over all baselines, demonstrating the effectiveness of differentiable physical constraint enforcement during inference. While GIMO attains the lowest GD, suggesting slightly better orientation alignment, our method maintains a comparable score while ensuring significantly fewer collisions and better endpoint accuracy. These results validate that our framework effectively balances geometric precision and physical feasibility, enabling reliable object trajectory generation in complex real-world scenes. An illustration is shown in Fig.~\ref{fig:hdepic_qualitative} and more in the supplementary.

\begin{table}[t!]
\centering
\small
\caption{\textbf{Cross-dataset evaluation.} Following the setup of ~\cite{yoshida2025generating}, we compare \methodname against the baselines on HOT3D dataset after training on Ego-Exo4D, thus demonstrating our superior performance on unseen scenes and cross-dataset generalization.}
\label{tab:model_comparison_hot3d}
\vspace{-2mm}
\begin{tabular}{lccc}
\toprule
\textbf{Model} & \textbf{ADE$\downarrow$} & \textbf{FDE$\downarrow$} & \textbf{GD$\downarrow$} \\
\midrule
% GMT & {0.230} & {0.023} & {1.415} \\
GIMO~\cite{zheng2022gimo} & \underline{0.299} & \underline{0.436} & {2.06} \\
CHOIS~\cite{li2024controllable} & {0.513} & {0.571} & {2.46} \\
SPOT~\cite{hsu2025spotse3posetrajectory} & {1.018} & {1.082} & {2.535} \\
DP3~\cite{Ze2024dp3} & {1.019} & {1.096} & {2.541} \\
M2Diffuser~\cite{yan2025m2diffuser} & {1.079} & {1.157} & {2.525} \\
Egoscaler~\cite{yoshida2025generating} & {0.351} & {0.540} & \textbf{0.856}\\
% \methodname w/o Guidance& \underline{0.285} & \textbf{0.052} & {1.62} \\
\methodname  & \textbf{0.265} & \textbf{0.027} & \underline{1.49} \\
\bottomrule
\end{tabular}
\vspace{-5mm}
\end{table}

%%%%%%%%%%%%%%%%%%%%
\subsection{Zero-shot Scenarios}
\label{subsec:egoexo4d}

\noindent\textbf{Datasets.}
Following the same setup as EgoScaler~\cite{yoshida2025generating}, we evaluate our framework under zero-shot conditions on two complementary egocentric datasets: Ego-Exo4D~\cite{grauman2024ego} and HOT3D~\cite{banerjee2025hot3d}. 
Ego-Exo4D is a large-scale multimodal benchmark capturing both egocentric and exocentric viewpoints of human activities, comprising $1{,}286$ hours of video from $740$ participants across $123$ real-world environments. 
We extract sequences involving explicit object motion and align them temporally to obtain dense 6DoF trajectories. 
HOT3D provides high-precision 3D ground-truth annotations for egocentric object motion, with approximately $833$ minutes of synchronized multi-view recordings from $19$ participants manipulating $33$ rigid objects. 
Following the preprocessing protocol of EgoScaler, we uniformly sample frames at $20$\,fps to ensure temporal consistency, resulting in $27{,}788$ training samples from Ego-Exo4D and $1{,}652$ test samples from HOT3D. 
Further dataset statistics and preprocessing details are provided in the supplementary material.

% \hz{Add more details in supplementary.}

\noindent\textbf{Results.} 
Tab.~\ref{tab:model_comparison_hot3d} summarizes the quantitative comparison on the HOT3D zero-shot test set. While Egoscaler~\cite{yoshida2025generating} attains the lowest GD, its position errors remain substantially higher, suggesting limited spatial consistency despite better orientation alignment. In contrast, our model achieves balanced performance across both translational and rotational metrics, confirming its ability to produce geometrically coherent and physically plausible 6DoF trajectories under unseen conditions, as qualitatively shown in Fig.~\ref{fig:hot3d_qualitative}.

% COMBINING INPUT CONDITION< AND GUIDANCE COST TABLES TO SAVE SPACE

% \begin{table*}[t]
% \centering
% \begin{tabular}{lccccc}
% \toprule
% \textbf{Model} & \textbf{ADE$\downarrow$} & \textbf{FDE$\downarrow$} & \textbf{Frechet Dist.$\downarrow$} & \textbf{Geodesic Dist.$\downarrow$} & \textbf{Collision Rate$\downarrow$} \\
% \midrule
% Ours w/o $\mathcal{P}$ & 0.305 & 0.110 & 0.205 & \textbf{1.121} & 2.9\% \\
% Ours w/o Action Label & 0.330 & 0.147 & 0.213 & 1.168 & 3.1\% \\
% Ours w/o $\mathbf{x}_T$ & 0.386 & 0.619 & 0.239 & 1.261 & 3.1\% \\
% Ours w/o History & 0.405 & 0.207 & 0.275 & 1.293 & 3.4\% \\
% \hdashline
% Ours w/o $\mathcal{J}_{\text{vel}}$ & 0.312 & 0.128 & 0.211 & 1.150 & 3.5\% \\
% Ours w/o $\mathcal{J}_{\text{rot}}$ & 0.312 & 0.128 & 0.211 & 1.150 & 3.6\% \\
% Ours w/o $\mathcal{J}_{\text{coll}}$ & {0.310} & {0.123} & {0.210} & 1.142 & 11.6\% \\
% \midrule
% Ours (Full) & \textbf{0.278} & \textbf{0.102} & \textbf{0.197} & {1.141} & \textbf{2.5\%} \\
% \bottomrule
% \end{tabular}
% \caption{Ablation study of model design and guidance components on HD-Epic}
% \label{tab:ablation_combined}
% \end{table*}

\begin{table}[t]
\caption{\textbf{Input and Guidance Ablation.} Ablation analysis on HD-EPIC assessing input conditioning modalities (top) and gradient guidance costs (bottom).}
\label{tab:ablation_combined}
\vspace{-2mm}
\centering
\small
\setlength{\tabcolsep}{3.6pt}
\begin{tabular}{p{1.9cm}ccccc}
\toprule
\textbf{Model} & \textbf{ADE$\downarrow$} & \textbf{FDE$\downarrow$} & \textbf{Frechet$\downarrow$} & \textbf{Geodesic$\downarrow$} & \textbf{Coll.$\downarrow$} \\
\midrule
w/o $\mathcal{P}$ & 0.305 & 0.110 & 0.205 & \textbf{1.121} & 2.9\% \\
w/o Action & 0.330 & 0.147 & 0.213 & 1.168 & 3.1\% \\
w/o $\mathbf{x}_T$ & 0.386 & 0.619 & 0.239 & 1.261 & 3.1\% \\
w/o History & 0.405 & 0.207 & 0.275 & 1.293 & 3.4\% \\
\hdashline
w/o $\mathcal{J}_{\text{vel}}$ & 0.312 & 0.128 & 0.211 & 1.150 & 3.5\% \\
w/o $\mathcal{J}_{\text{rot}}$ & 0.312 & 0.128 & 0.211 & 1.150 & 3.6\% \\
w/o $\mathcal{J}_{\text{coll}}$ & {0.310} & {0.123} & {0.210} & 1.142 & 11.6\% \\
\midrule
Ours (Full) & \textbf{0.278} & \textbf{0.102} & \textbf{0.197} & {1.141} & \textbf{2.5\%} \\
\bottomrule
\end{tabular}
\end{table}

\begin{table}[t]
\centering
\small
\caption{\textbf{Architecture Ablation.} Study of different layer configurations (Mamba-Transformer-Mamba layers) on HD-EPIC.}
\label{tab:ablation_architecture}
\vspace{-2mm}
\setlength{\tabcolsep}{3.6pt}
\begin{tabular}{p{2.0cm}ccccc}
\toprule
% \textbf{Layer Config (M-T-M)} & \textbf{ADE$\downarrow$} & \textbf{FDE$\downarrow$} & \textbf{Frechet$\downarrow$} & \textbf{Geodesic$\downarrow$} & \textbf{Coll.$\downarrow$} 
\textbf{M-T-M Config} & \textbf{ADE$\downarrow$} & \textbf{FDE$\downarrow$} & \textbf{Frechet$\downarrow$} & \textbf{Geodesic$\downarrow$} & \textbf{Coll.$\downarrow$} \\
\midrule
0-12-0 & 0.440 & \textbf{0.072} & 0.292 & 1.271 & \underline{2.9\%} \\
1-10-1 & 0.312 & 0.108 & 0.211 & \textbf{1.138} & 3.1\% \\
5-2-5 & \underline{0.305} & 0.113 & 0.207 & 1.164 & 3.7\% \\
6-0-6 & 0.327 & 0.137 & \underline{0.206} & 1.149 & 4.6\% \\
\midrule
Ours (3-6-3) & \textbf{0.279} & \underline{0.102} & \textbf{0.197} & \underline{1.141} & \textbf{2.5\%} \\
\bottomrule
\end{tabular}
\vspace{-5mm}
\end{table}

% \todo{wait for abhishek's results to analyze}
%%%%%%%%%%%%%%%%%%%%%
\subsection{Ablation Study}
\label{subsec:ablations}
We conduct an ablation study to validate the effectiveness of various input conditioning and guidance components. As shown in Tab.~\ref{tab:ablation_combined}, removing any individual component consistently degrades performance across most evaluation metrics, confirming their complementary roles. 

\noindent\textbf{Input Conditioning.}
Excluding the scene point cloud or end position guidance leads to the largest increase in ADE/FDE, indicating that accurate spatial grounding and goal conditioning are crucial for reliable motion generation. Similarly, removing history inputs significantly harms temporal coherence, resulting in higher geodesic and Frechet distances. The action label also provides beneficial semantic context, helping the model disambiguate motion intent.

\noindent\textbf{Guidance Cost Functions.}
Among the guidance terms, eliminating velocity smoothness or rotation consistency slightly increases trajectory deviation and collision rates, demonstrating their contribution to producing physically plausible and stable motions. Notably, omitting collision avoidance results in a sharp rise in collision rate (from 2.5\% → 11.6\%), implying that incorporating this term reduces collisions by about 79\%, thus emphasizing its importance for generating physically feasible trajectories.

\noindent\textbf{Layer Configurations.}
In Tab.~\ref{tab:ablation_architecture}, we investigate the impact of layer distribution across the three stages of our Motion Generation Model. We compare hybrid Mamba-Transformer configurations against pure architectures: a Transformer-only model (0-12-0) and a Mamba-only model (6-0-6). Our balanced 3-6-3 configuration achieves the best overall performance across all metrics.

\section{Conclusion}
\vspace{-0.1cm}
In this work, we introduced EgoFlow, a generative framework that unifies flow matching in $\mathbb{R}^{9}$ with gradient-guided inference for physically grounded egocentric trajectory generation. By integrating long-horizon temporal modeling through a hybrid Mamba–Transformer–Perceiver architecture and embedding differentiable physical priors directly into the sampling process, EgoFlow achieves a rare balance between realism, controllability, and efficiency.

Extensive experiments across egocentric benchmarks demonstrate that EgoFlow consistently outperforms diffusion-based and transformer-only baselines in both spatial accuracy and physical plausibility, while generalizing effectively to unseen domains without fine-tuning. The results suggest that flow-based motion generation can serve as a strong alternative to diffusion models for continuous 3D prediction tasks.

Looking forward, we see EgoFlow as a step toward holistic embodied scene understanding, where perception, motion, and interaction are modeled in a unified generative framework. Future directions include extending our approach to deformable objects, integrating closed-loop sensory feedback, and coupling EgoFlow with robot policy learning to enable goal-directed, physically consistent action generation. %from first-person perception.

\noindent\textbf{Limitations}
Our current formulation handles diverse objects including semi-deformable items (clothes, paper), but doesn't model volumetric deformations or shape changes during its motion. We also assume static environments with known geometry, which limits applicability in scenes with moving agents or dynamic clutter. Addressing these limitations presents promising directions for future work.  
% Future extensions could incorporate explicit deformation dynamics and dynamic scenes.

\noindent\textbf{Acknowledgments}
This research was partially funded by the German Federal Ministry of Education and Research through the ExperTeam4KI funding program for UDance (Grant No.\ 16IS24064).
{
    \small
    \bibliographystyle{ieeenat_fullname}
    \bibliography{main}
}

\clearpage
\setcounter{page}{1}
\maketitlesupplementary

%%%%%%%%%%%%%%%%%%%%%%%
% catalogue
% 1. architectural hyperparameters of our framework, including layer configurations, hidden dimensions, and attention head counts \as{DONE}

% %
% 2. Effect of optimization steps Sec.~\ref{sec:supp_optimize_steps} (computational cost, metrics) \as{DONE}

% %
% 3. dataset preprocessing details (rewrite Egoscaler)

% %
% 4. diffusion-based and flow-matching  \as{DONE, add figures quali}

% %
% 5. visualize representative frames from HD-EPIC pseudo gt \as{DONE}

% %
% 6. More qualitative results (1 successful and failure case each dataset)
%%%%%%%%%%%%%%%%%%%%%%%

In this supplementary material, we provide additional details and analyses to complement the main paper. The supplementary is organized as follows:

First, in Sec.~\ref{sec:supp_data_preprocess}, we describe the dataset preprocessing steps to convert raw data into a consistent format for training and evaluation. 
In Sec.~\ref{sec:supp_model_arch}, we provide architectural hyperparameters of our framework, including layer configurations, hidden dimensions, and attention head counts. 
Next, we further analyze the effect of optimization steps during inference in Sec.~\ref{sec:supp_optimize_steps}, comparison of different generation paradigms in Sec.~\ref{sec:supp_diffusion_vs_flow}, and the impact of goal conditioning in Sec.~\ref{sec:supp_goal_conditioning}.

Finally, a proof-of-concept application of our framework for robot manipulation is presented in Sec.~\ref{sec:supp_application}.

%%%%%%%%%%%%%%%%%%%%%%
\section{Dataset Preprocessing}
\label{sec:supp_data_preprocess}
In this section, we provide more details on how we preprocess the datasets used in our experiments into a consistent format.

\noindent\textbf{HD-EPIC~\cite{perrett2025hd}.} The HD-EPIC dataset lacks explicit annotations such as object bounding boxes or semantic labels. To compensate for this, large static elements (e.g., tables, drawers) are manually modeled in Blender and aligned with the scene's point cloud to define fixed bounding boxes. For smaller, movable items like coffee makers and knives, the dataset includes temporal segments marking object motion, along with 2D segmentation masks and estimated 3D centers. Using these cues, we synchronize timestamps with SLAM outputs and extract sparse 2D-3D correspondences via MPS data captured by Aria glasses. Monocular depth maps are then inferred with DepthAnythingv2~\cite{yang2024depth}, and depth values are linearly scaled using the correspondences to recover metric scale. Finally, we reconstruct each object's 3D bounding box.\\

\noindent\textbf{HD-EPIC Validation on ADT~\cite{Pan_2023_ICCV}.} 
% \as{TODO, details and link with videos and pics}
Since the HD-EPIC dataset provides only sparse annotations of object positions, we developed an algorithm to generate dense annotations of object motion, as described in Sec.~\ref{subsec:hdepic}. To validate the effectiveness of this algorithm, we applied it to the Aria Digital Twin dataset, which offers dense ground-truth trajectories for dynamic objects across their motion sequences. Table~\ref{tab:adt_stats} summarizes the evaluation results on this synthetic dataset. As shown, for sufficiently long trajectories, the 3D Euclidean error between the ground-truth and the predicted object positions remains minimal. Nevertheless, we exclude this dataset from our main experiments, as its limited size and lack of real-world diversity prevent it from providing meaningful training data for generative models.

\begin{table}[h]
\centering
\small
\setlength{\tabcolsep}{4pt}
\begin{tabular}{cp{1.3cm}p{1.1cm}p{1.2cm}p{1.2cm}}
\toprule
\textbf{\#Traj.} & \textbf{ADE Mean (cm)} & \textbf{ADE Median (cm)} & \textbf{Traj. Length Mean (m)} & \textbf{Traj. Length Median (m)} \\
\midrule
505 & 20.35 & 15.76 & 3.90 & 3.06 \\
\bottomrule
\end{tabular}
\caption{\textbf{ADT Statistics.} We study the effectiveness of our object position estimation approach and verify it on the ADT dataset.}
\label{tab:adt_stats}
\end{table}

\begin{figure*}[h]
    \centering
    \includegraphics[width=\linewidth]{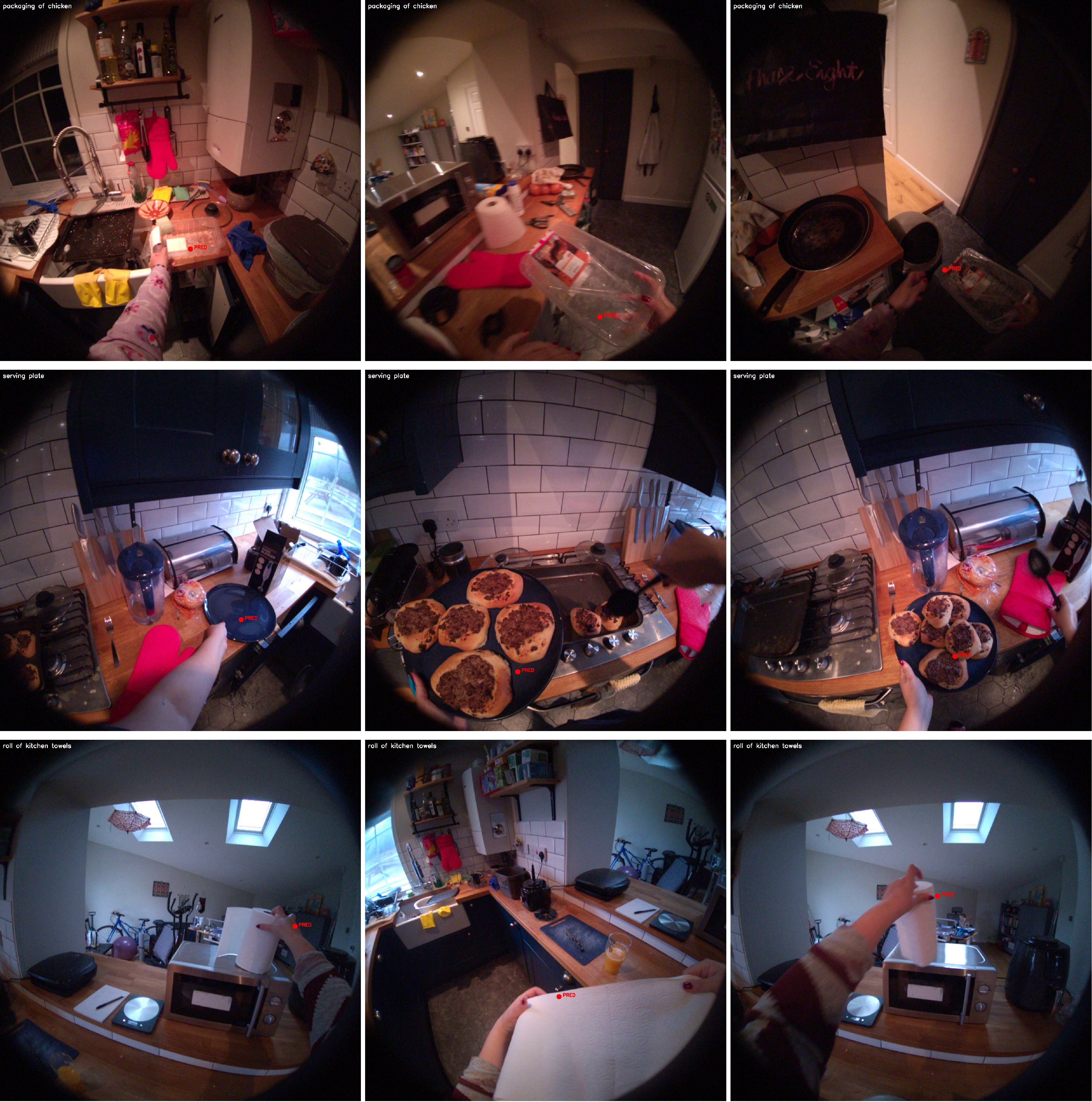}
    \caption{We project our object position calculated by our object position estimation algorithm using the hand poses of MPS as described in Sec. \ref{subsec:hdepic} onto the egocentric video frames to show the correctness and accuracy of our approach.}
    \label{fig:verify-hdepic}
\end{figure*}

\begin{figure*}[h]
    \centering
    \includegraphics[width=\linewidth]{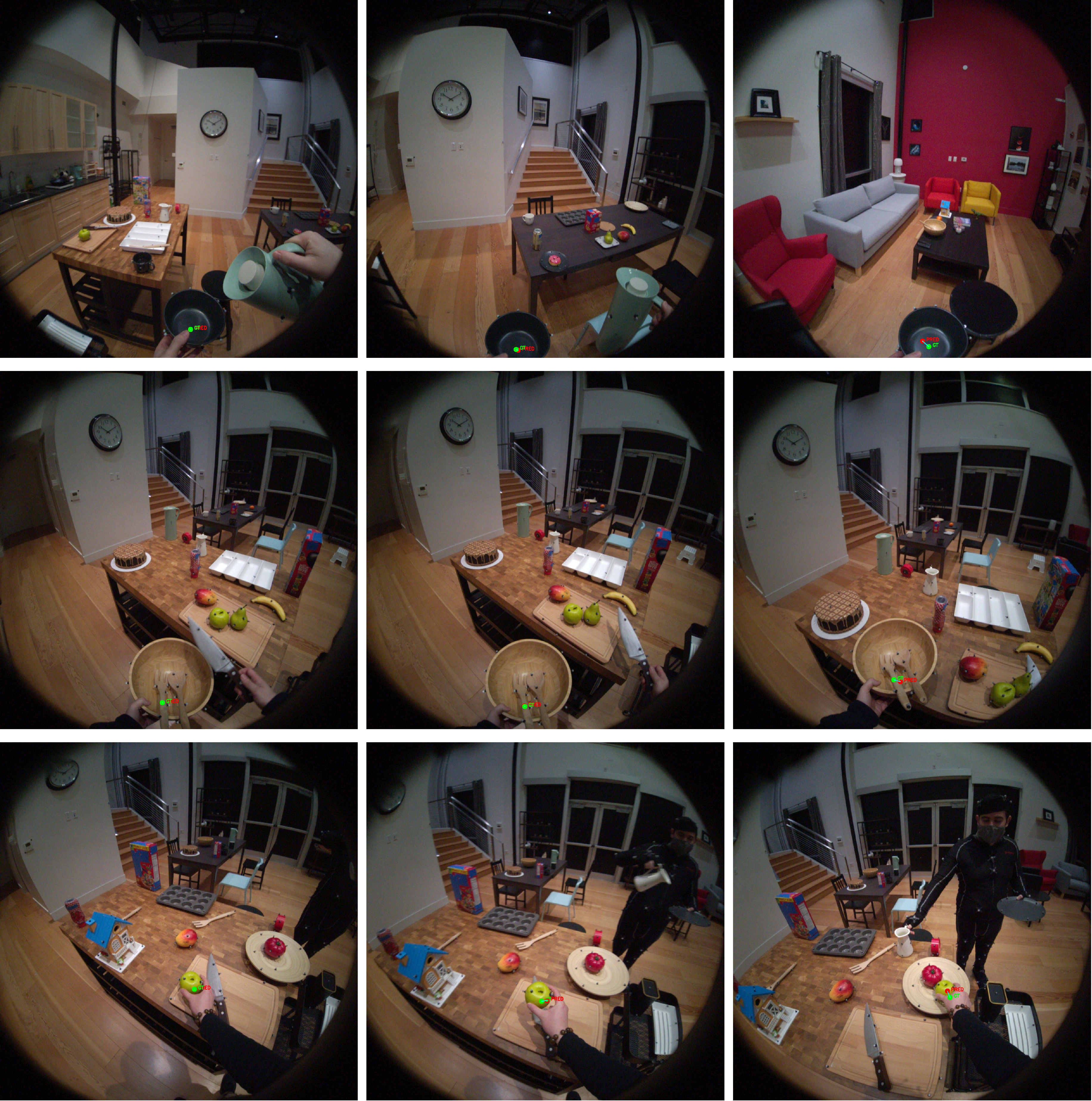}
    \caption{We plot our object position calculation algorithm on the ADT dataset. Since ADT has rich annotations, it works as an ideal demonstration of the effectiveness of our algorithm to generate dense object motions on the HD-EPIC dataset.}
    \label{fig:verify-hdepic-adt}
\end{figure*}

Additionally, we project the computed 3D object locations during their motion onto egocentric video frames to verify the algorithm's effectiveness. We demonstrate this in Fig. \ref{fig:verify-hdepic} and Fig. \ref{fig:verify-hdepic-adt}. We also provide supporting videos of these object motions for a more holistic evaluation. \\

\noindent\textbf{Ego-Exo4D~\cite{grauman2024ego} \& HOT3D~\cite{banerjee2025hot3d}.} 
For the Ego-Exo4D dataset, egocentric video streams were divided into short clips of approximately four seconds centered on annotated timestamps. Segments involving explicit object-hand interactions were automatically selected.
An open-vocabulary segmentation model (Grounded-SAM~\cite{ren2024grounded}) was employed to identify the manipulated object in the initial frame, and the object was subsequently tracked throughout the clip using SpaTracker~\cite{xiao2024spatialtracker}.
Depth maps were estimated for all frames using the DepthAnythingv2~\cite{yang2024depth}, and RGB-D images were converted into point clouds.

For the HOT3D dataset, ground-truth 6DoF object trajectories captured with OptiTrack infrared cameras were utilized directly.
Since temporal boundaries and textual descriptions were not provided in the original dataset, videos were divided into four second clips and action intervals were automatically localized using GPT-4o-assisted temporal annotation. Depth maps were further estimated to align all trajectories with the same coordinate convention as the Ego-Exo4D data.

%%%%%%%%%%%%%%%%%%%%%%

\section{Architecture Details}
\label{sec:supp_model_arch}
We provide more details on the architecture hyperparameters of our proposed model in Tab. \ref{tab:hyperparams}.
\begin{table}[h]
  \centering
  \small
  \begin{tabular}{lc}
  \toprule
  \textbf{Component} & \textbf{Value} \\
  \midrule
  \multicolumn{2}{l}{\textit{Core Architecture}} \\
  Hidden dimension & 768 \\
  Dropout & 0.1 \\
  \addlinespace
  \multicolumn{2}{l}{\textit{Stage 1: Bidirectional Mamba}} \\
  Number of layers & 3 \\
  SSM state dimension & 256 \\
  \addlinespace
  \multicolumn{2}{l}{\textit{Stage 2: Transformer}} \\
  Number of layers & 6 \\
  Attention heads & 12 \\
  \addlinespace
  \multicolumn{2}{l}{\textit{Stage 3: Bidirectional Mamba}} \\
  Number of layers & 3 \\
  SSM state dimension & 256 \\
  \addlinespace
  \multicolumn{2}{l}{\textit{Conditioning Encoders}} \\
  Scene encoder (PointNet++) & 512 \\
  Motion encoder & 256 \\
  Bounding box encoder & 256 \\
  \quad BBox attention heads & 4 \\
  Category encoder (CLIP ViT-B/32) & 256 \\
  Semantic bbox encoder & 256 \\
  Goal pose encoder & 256 \\
  Motion-bbox Perceiver latent & 256 \\
  \quad Perceiver attention heads & 4 \\
  \quad Perceiver attention layers & 4 \\
  \midrule
  Total conditioning dimension & 1792 \\
  \bottomrule
  \end{tabular}
  \caption{Model architecture hyperparameters.}
  \label{tab:hyperparams}
  \end{table}

 The model uses 768-dimensional hidden representations across all stages with 0.1 dropout. Stage 1 contains 3 bidirectional Mamba layers with SSM state dimension 256. Stage 2 contains 6 Transformer layers with 12 attention heads per layer.
  Stage 3 contains 3 bidirectional Mamba layers with SSM state dimension 256.

  The conditioning pipeline consists of: PointNet++ producing 512-dimensional scene features, motion encoder producing 256 dimensions, bounding box encoder with 4-head attention producing 256 dimensions, CLIP ViT-B/32 producing 256-dimensional
   category embeddings, semantic bbox encoder producing 256 dimensions, goal pose encoder producing 256 dimensions, and a Perceiver module with 256 latent dimensions, 4 attention heads, and 4 layers. The concatenated conditioning vector is
  1792 dimensions.
%%%%%%%%%%%%%%%%%%%%%%%

\section{Analysis of Optimization Steps}
\label{sec:supp_optimize_steps}
We analyze the trade-off between guidance optimization and inference efficiency (Tab. \ref{tab:ablation_opt_steps} and Fig. \ref{fig:opt_steps}). Test-time optimization primarily affects collision avoidance: collision rate decreases from 11.9\% (0 steps) to 2.5\% (50 steps), while trajectory errors remain largely stable (ADE: 0.273-0.279m, FDE: 0.102-0.119m). Noticeably, ADE gets slightly worse in this case, as the model is often encouraged to slightly deviate from its path to produce more collision-free trajectories, and hence it goes through more unoccupied space to avoid other objects, while noticeably reaching its destination. 
Inference time scales approximately linearly from 0.254s to 0.480s per trajectory. We use 50 steps for our experiments to ensure collision-free generation.

\begin{table}[t]
\centering
\small
\setlength{\tabcolsep}{3.4pt}
\begin{tabular}{p{0.61cm}cccccc}
\toprule
\textbf{Steps} & \textbf{ADE$\downarrow$} & \textbf{FDE$\downarrow$} & \textbf{Geodesic$\downarrow$} & \textbf{Coll.$\downarrow$} & \textbf{Time(s)$\downarrow$} \\
\midrule
% 0 & \textbf{0.273} & \underline{0.119} & \textbf{0.194} & \underline{1.151} & 11.9\% & \textbf{0.254} \\
1 & \textbf{0.273} & \underline{0.119} &  \underline{1.151} & 11.7\% & \underline{0.259} \\
5 & \textbf{0.273} & \underline{0.119} &  \underline{1.151} & 11.0\% & 0.277 \\
10 & \textbf{0.273} & \underline{0.119} &  \underline{1.151} & 10.0\% & 0.301 \\
20 & \textbf{0.273} & \underline{0.119} & \underline{1.151} & 8.2\% & 0.345 \\
30 & \textbf{0.273} & 0.120 &  \underline{1.151} & 6.7\% & 0.391 \\
40 & \textbf{0.273} & 0.120 &  \underline{1.151} & \underline{4.6\%} & 0.436 \\
50 & 0.279 & \textbf{0.102} & \textbf{1.141} & \textbf{2.5\%} & 0.480 \\
\bottomrule
\end{tabular}
\caption{Ablation study of optimization steps on HD-EPIC. Time denotes inference time per trajectory in seconds.}
\label{tab:ablation_opt_steps}
\end{table}

\pgfplotsset{compat=1.18}

\begin{figure}[t]
\centering
\begin{tikzpicture}
\pgfplotsset{set layers}
\begin{axis}[
    width=0.95\columnwidth,
    height=7cm,
    xlabel={Optimization Steps},
    ylabel={Error (m)},
    ylabel style={black},
    legend style={at={(0.5,1.08)}, anchor=south, legend columns=3, font=\small},
    grid=major,
    grid style={dashed,gray!30},
    mark size=3pt,
    line width=1.2pt,
    axis y line*=left,
    ymin=0.09, ymax=0.29,
    name=mainplot,
]

% ADE curve (red)
\addplot[color=red, mark=*, thick] coordinates {
    (0, 0.273) (1, 0.273) (5, 0.273) (10, 0.273)
    (20, 0.273) (30, 0.273) (40, 0.273) (50, 0.279)
};
\addlegendentry{ADE}

% FDE curve (blue)
\addplot[color=blue, mark=square*, thick] coordinates {
    (0, 0.119) (1, 0.119) (5, 0.119) (10, 0.119)
    (20, 0.119) (30, 0.120) (40, 0.120) (50, 0.102)
};
\addlegendentry{FDE}

% Add legend entry for Collision Rate from second axis
\addlegendimage{color=green!60!black, mark=triangle*, thick}
\addlegendentry{Collision Rate}

\end{axis}

% Secondary y-axis for Collision Rate
\begin{axis}[
    width=0.95\columnwidth,
    height=7cm,
    xlabel={},
    ylabel={Collision Rate (\%)},
    ylabel style={green!60!black},
    legend=false,
    mark size=3pt,
    line width=1.2pt,
    axis y line*=right,
    axis x line=none,
    ymin=0, ymax=14,
]

% Collision Rate curve (green)
\addplot[color=green!60!black, mark=triangle*, thick] coordinates {
    (0, 11.9) (1, 11.7) (5, 11.0) (10, 10.0)
    (20, 8.2) (30, 6.7) (40, 5.6) (50, 2.5)
};

\end{axis}
\end{tikzpicture}
\caption{Effect of optimization steps on motion quality metrics.}
\label{fig:opt_steps}
\end{figure}
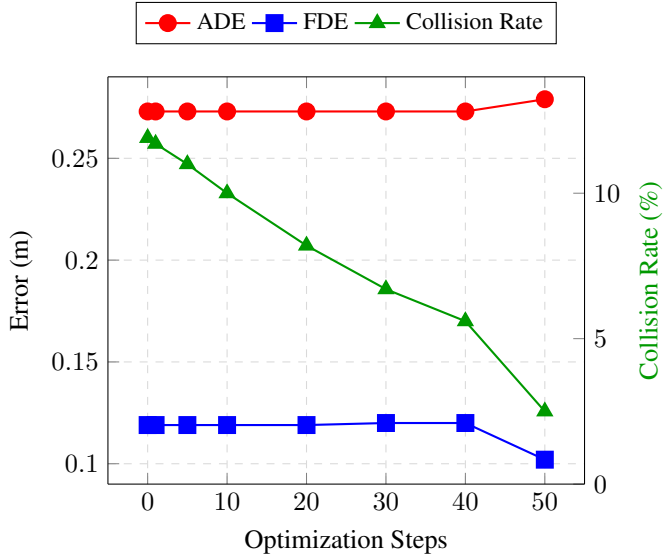

%%%%%%%%%%%%%%%%%%%%%%%%%%%%%%%%%%%
\section{Analysis of Generation Paradigms}
\label{sec:supp_diffusion_vs_flow}
\begin{figure*}[h]
    \centering
    \includegraphics[width=\linewidth]{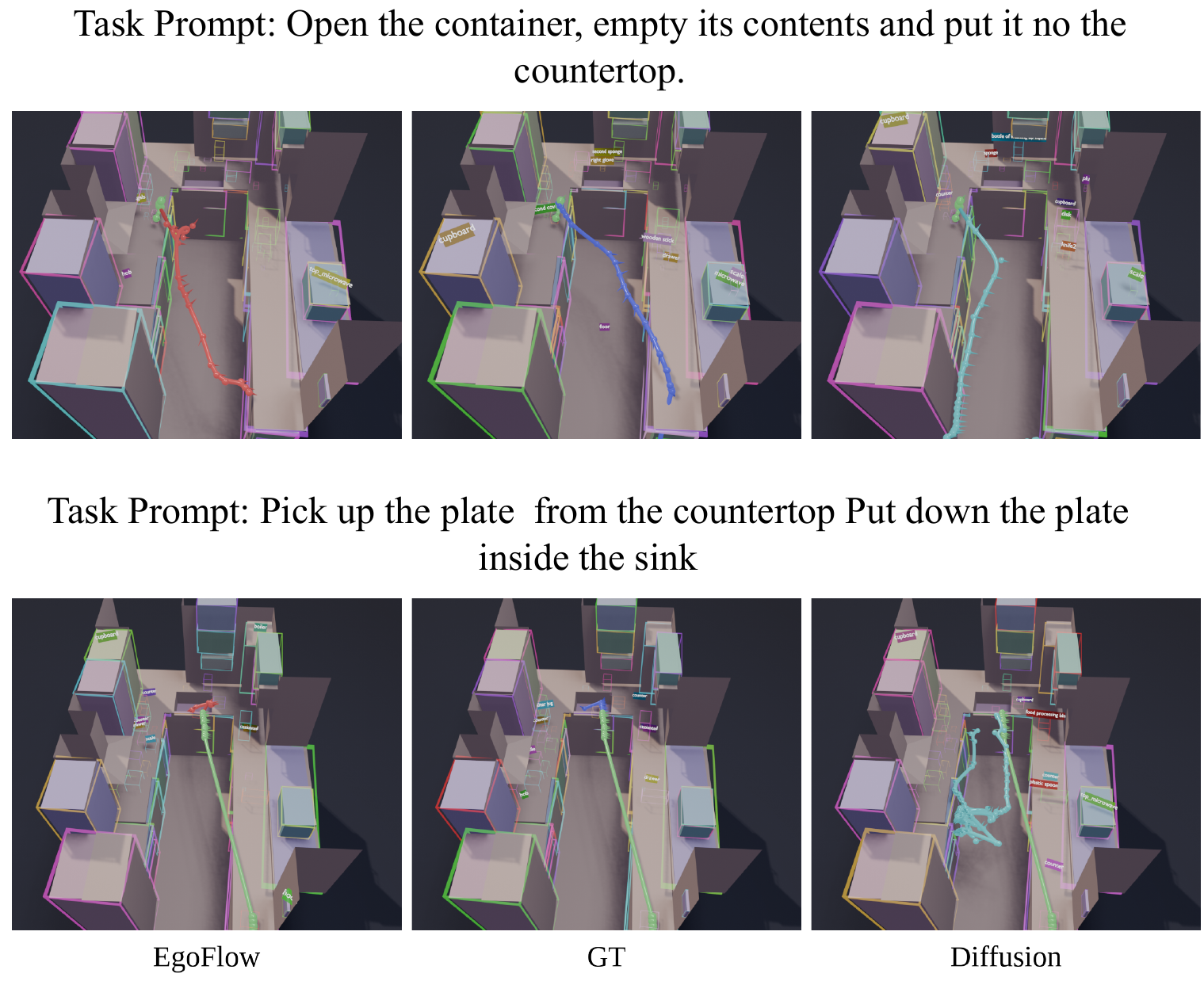}
    \caption{\textbf{Flow Matching vs Diffusion.} We show a qualitative comparison of using flow matching vs a diffusion paradigm while using the same model as the backbone. We observe that flow matching is more faithful to the ground truth and conditionings due to its simpler objective and more straight-line paths from noise to data distribution over the complex denoising steps of its diffusion counterpart. The green part of the trajectory shows the history, while Red, Blue and Cyan depict the flow matching, ground truth, and the diffusion-based predictions respectively.}
    \label{fig:diffusion-vs-flow}
\end{figure*}

We compare our flow matching approach against diffusion-based generation using the same architecture and training data (Tab.~\ref{tab:diffusion_vs_flow}). Flow matching achieves substantially better trajectory quality (ADE: 0.279m vs 0.658m, FDE: 0.102m vs 0.549m) and is nearly an order of magnitude  faster. When applying guidance to diffusion, we observe a fundamental trade-off: collision rate drops to nearly 1\% but trajectory errors worsen (ADE: 0.692m, FDE: 0.660m) and inference time doubles to 5.561s. This degradation occurs because guidance is applied at every denoising step, causing the model to overemphasize collision avoidance and reroute trajectories into empty regions rather than toward task-relevant endpoints. In contrast, flow matching's deterministic straight-path interpolation allows guidance to refine constraints without derailing the overall motion plan. These results motivate our choice of flow matching for physically plausible object motion generation where both accuracy and constraint satisfaction are critical. Fig.~\ref{fig:diffusion-vs-flow} further illustrates the qualitative differences, showing that flow matching produces trajectories more faithful to the ground truth and conditioning.

\begin{table*}[t]
\centering
\small
\setlength{\tabcolsep}{4pt}
\begin{tabular}{lcccccc}
\toprule
\textbf{Model} & \textbf{ADE$\downarrow$} & \textbf{FDE$\downarrow$} & \textbf{Geodesic$\downarrow$} & \textbf{Coll.$\downarrow$} & \textbf{Time(s)$\downarrow$} \\
\midrule
Diffusion w/o Guidance & 0.658 & 0.549 & 1.437 & 25.0\% & 2.745 \\
Diffusion & \underline{0.692} & \underline{0.660} & \underline{1.483} & \textbf{0.96\%} & 5.561 \\
\midrule
Ours (Flow Matching) & \textbf{0.279} & \textbf{0.102} & \textbf{1.141} & \underline{2.5\%} & \textbf{0.480} \\
\bottomrule
\end{tabular}
\caption{\textbf{Diffusion vs Flow Matching.} Comparison of diffusion-based and flow matching approaches on HD-EPIC. Time denotes inference time per trajectory in seconds.}
\label{tab:diffusion_vs_flow}
\end{table*}

%%%%%%%%%%%%%%%%%%%%%%

\section{Analysis of Goal Conditioning}
\label{sec:supp_goal_conditioning}
\begin{figure*}[h]
    \centering
    \includegraphics[width=0.9\linewidth]{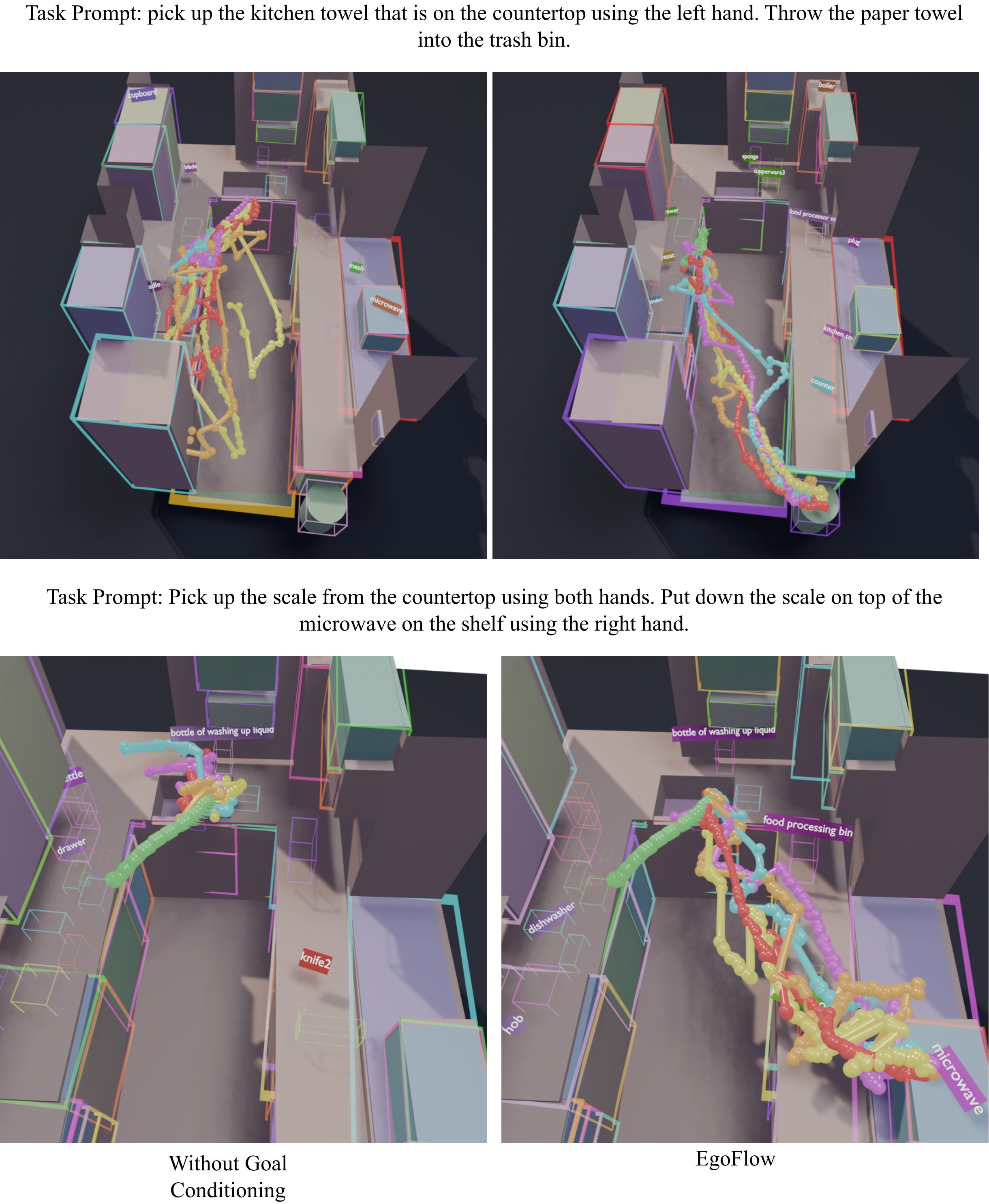}
    \caption{\textbf{Analysis of Goal Conditioning} We show a qualitative comparison of the effect of having goal conditioning as input to the model. The green part of the trajectory shows the history, while the other colors such as red, orange, cyan, yellow and purple show the \textbf{multiple possible paths} the object motion could take based on the history to the end goal.}
    \label{fig:ablation-goal-condition}
\end{figure*}

Here, we show the qualitative comparison in Fig.~\ref{fig:ablation-goal-condition} for the effect of having the end goal conditioned to the model as an input. Additionally, we show the \textbf{multiple possible paths} that the object could take based on the history and the conditioning. Supporting the quantitative results in Sec.~\ref{subsec:ablations}, we show that the ADE and FDE are worsened as the object fails to reach its end goal. 

%%%%%%%%%%%%%%%%%%%%%%%%

\section{Application}
\label{sec:supp_application}
\begin{figure*}[h]
    \centering
    \includegraphics[width=0.8\linewidth]{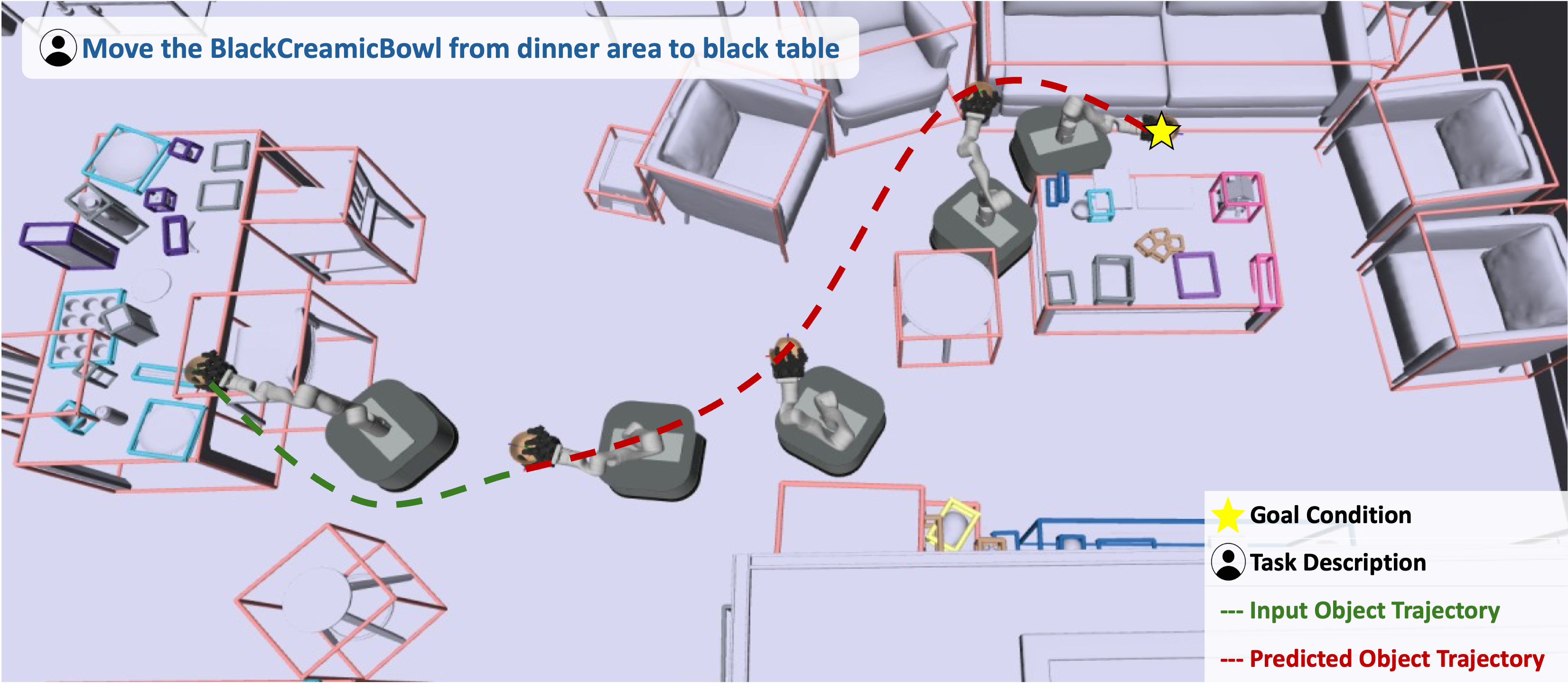}
    \caption{Visualization of robot mobile manipulation using  object trajectories. The robot follows the object paths while maintaining a fixed grasp pose.}
    \label{fig:supp_application}
\end{figure*}

As a proof-of-concept, we further explored the potential of our framework in real-world applications. Given predicted object trajectories, we employed inverse kinematics (IK) to generate robot manipulator motions that follow the planned object paths. We utilized the Mink solver~\cite{Zakka_Mink_Python_inverse_2025} on Mujuco~\cite{todorov2012mujoco} to compute joint configurations for Tidybot~\cite{wu2023tidybot} with Leap hand~\cite{shaw2023leap}, ensuring that the end-effector maintained a fixed grasp pose relative to the object throughout the manipulation. For visualization, we rendered the object trajectories with a mobile robot in Mujuco, as shown in \cref{fig:supp_application}.

It is worth noting that the current demonstrations still have several limitations. 
First, we assume that the grasp pose is known and remains fixed during manipulation, which does not always hold in real-world scenarios. 
Shifts in the grasp pose can lead to infeasibility in IK solutions. 
Second, dynamic feasibility is not explicitly enforced for high DoF robots such as humanoids or quadrupeds, which may cause balance loss or physically implausible motions during execution. 
Third, our current framework focuses on object-level collision avoidance and does not incorporate base motion planning, which is essential for achieving coordinated whole-body control. 
Overall, successful mobile manipulation requires the integration of grasp pose estimation, base motion planning, and dynamic feasibility constraints, which we leave as future work.

% {
%     \small
%     \bibliographystyle{ieeenat_fullname}
%     \bibliography{supp}
% }

\end{document}